%% file: main.tex
\pdfoutput=1

\documentclass[letterpaper]{article} 
\usepackage{aaai23}  
\usepackage{times}  
\usepackage{helvet}  
\usepackage{courier}  
\usepackage[hyphens]{url}  
\usepackage{graphicx} 
\usepackage{natbib}  
\usepackage{caption} 
\usepackage{latexsym}

\usepackage[T1]{fontenc}

\usepackage[utf8]{inputenc}

\usepackage[normalem]{ulem}

\usepackage{booktabs}

\usepackage{microtype}

\usepackage{multirow}

\usepackage{amsmath}
\usepackage{amsfonts}

\title{Understanding Text Classification Data and Models\\ Using Aggregated Input Salience}

\author {
    Sebastian Ebert,
    Alice Shoshana Jakobovits,
    Katja Filippova
}
\affiliations {
    Google Research\\
    \texttt{\{eberts,jakobovits,katjaf\}@google.com}
}

\begin{document}
\maketitle

\begin{abstract}
Realizing when a model is right for a wrong reason is not trivial and requires a significant effort by model developers.
In some cases an input salience method, which highlights the most important parts of the input, may reveal problematic reasoning. But scrutinizing highlights over many data instances is tedious and often infeasible.
Furthermore, analyzing examples in isolation does not reveal general patterns in the data or in the model's behavior.
In this paper we aim to address these issues and go from understanding single examples to understanding entire datasets and models.
The methodology we propose is based on aggregated salience maps, to which we apply clustering, nearest neighbor search and visualizations.
Using this methodology we address multiple distinct but common model developer needs by showing how problematic data and model behavior can be identified and explained -- a necessary first step for improving the model.\\
\textbf{Warning: Due to the usage of a toxicity dataset this paper contains content that may be offensive or upsetting.}
\end{abstract}

\section{Introduction}

\begin{figure}[t!]
    \centering
    \includegraphics[width=\columnwidth]{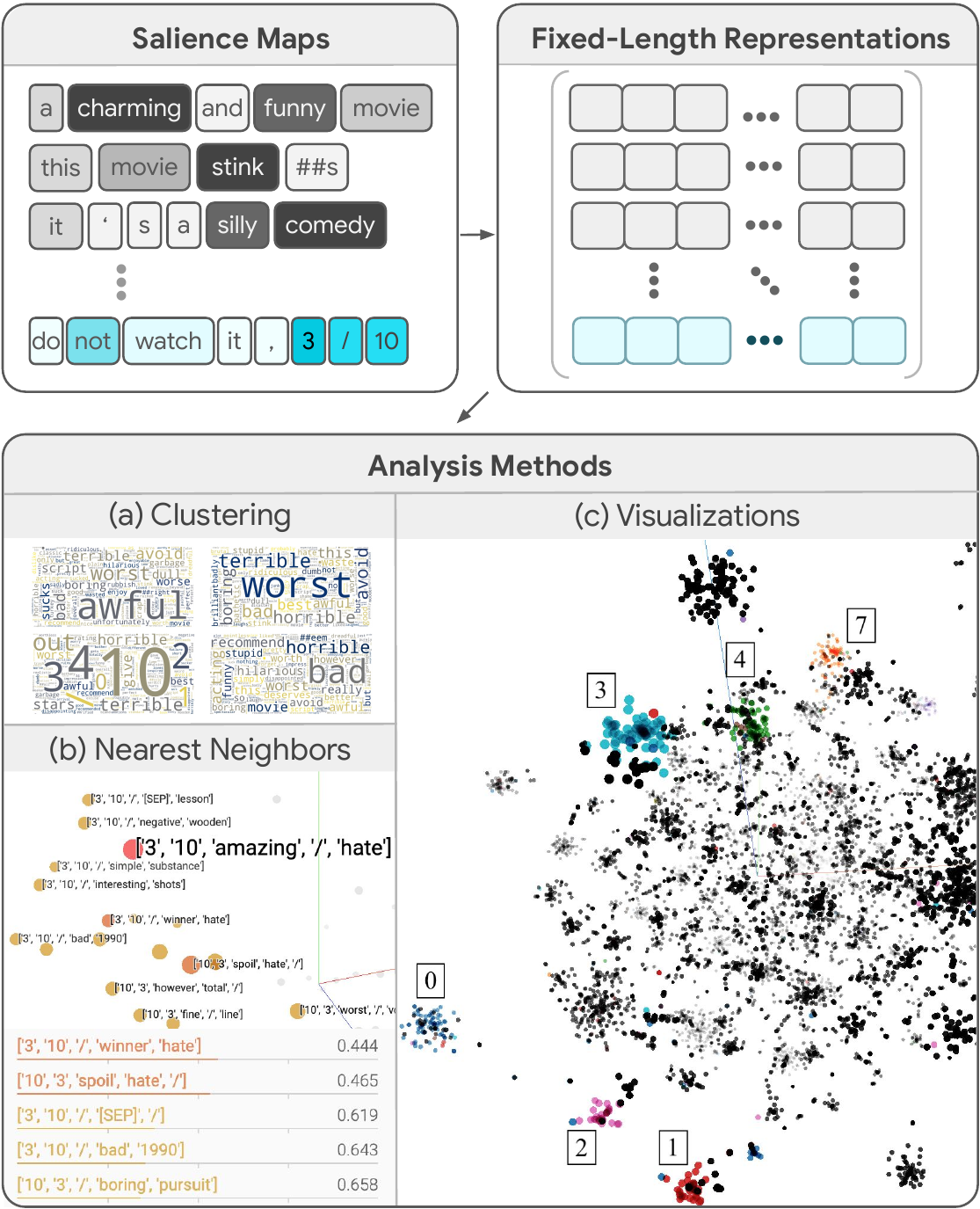}
    \caption{An overview of the proposed method: from raw input \textbf{salience maps}, compute \textbf{fixed-length representations}, and apply \textbf{analysis methods} such as clustering, nearest neighbors search and visualizations.
    Showing some examples from the IMDB dataset, the high-salience tokens `\textit{3}', `\textit{/}' and `\textit{10}' in the last example's salience map are reflected in the analyses:
    (a) Word clouds computed from four clusters show that one cluster is dominated by number tokens. 
    (b) Nearest neighbors of the example reveal more examples with similar high-salience token combinations.
    (c) A t-SNE visualization of the IMDB test set, where the turquoise point cloud labeled `\textit{3}' contains all examples similar to the original.
    }
    \label{fig:overview}
\end{figure}

Deploying ML-powered models requires confidence in their reliability. While strong performance on an evaluation set is a prerequisite, it is not sufficient since it may hide poor generalization patterns. It is the responsibility of the model developer to analyze the model and proactively search for problematic patterns, discover its sensitivities, or uncover sources of erroneous predictions. 

Over the past few years, the explainable AI community has developed many methods and created platforms that support developers in debugging their models \cite[inter alia]{NorJenKoc+19v,tenney-etal-2020-language}.
But most efforts have been devoted to analyzing \emph{single} predictions, often with the help of input salience \cite[inter alia]{li-etal-2016-visualizing, montavon-lrp-overview-2019} or training data attribution techniques \cite{koh-liang-if-2017,pruthi-tracin-20}.
Though already somewhat useful \cite{lertvittayakumjorn-toni-2021-explanation}, single-point explanations
may lead the developer to discover false generalizations if the analyzed examples are unrepresentative.
For instance, when predicting a contradiction in Natural Language Inference tasks \cite{bowman-etal-2015-nli}, does a very high salience weight on the word `\textit{not}', imply that the model learned an overly simplistic pattern and consistently ignores the context when `\textit{not}' is present?
Understanding whether shallow reasoning is applied systematically would require going through a large number of inputs, which is tedious and time consuming, rendering it virtually impossible to discover patterns learned from the whole dataset.
Conversely, patterns unintuitive for a human may be difficult to spot from individual examples even when the salience maps hints at its presence. 

In order to gather stronger support that the model is ``right for the right reasons'' we propose to shift focus from single point analyses to systematic patterns.
We propose to use standard input salience methods, convert the maps they produce into fixed-length representations, and analyze the resulting space using clustering, nearest neighbor search or interactive visualizations (Fig.~\ref{fig:overview}).
We use salience methods as opposed to corpus statistics \cite{gardner-etal-2021-competency} because they point at what is important \textit{for the model} \cite{bastings-filippova-2020-elephant} and because not every feature-label correlation results in a spurious correlation picked by a model \cite{eisenstein-2022-informativeness}.

With this aggregated view we are able to surface shallow reasoning patterns \cite{mccoy-etal-2019-right,rosenman-etal-2020-exposing} and data artifacts that the model is sensitive to \cite{gururangan-etal-2018-annotation,geva-etal-2019-modeling}.
This can be a first step towards correcting or augmenting the data and improving the model.
Furthermore, independently of whether a learned pattern poses a robustness threat, we show how the same techniques can help better explore and understand the characteristics of a dataset which we believe also falls under a developer's responsibility.

Addressing distinct but common developer needs, we make the following contributions:
\begin{enumerate}
    \item In Sec.~\ref{sec:problematic_patterns}, we demonstrate how clustering salience maps helps identify patterns in a dataset which result in a spurious correlation \cite{geirhos-2020-shortcuts} and reveal weaknesses of a dataset. Furthermore, we present a qualitative analysis enabled by clustering subsamples of the data which results in a more fine-grained characterization of the relationship between certain tokens and prediction than has been previously done.
    \item In Sec.~\ref{sec:model_sensitivity}, we describe a procedure grounded in salience clusters which facilitates finding out what a model is sensitive to. 
    \item In Sec.~\ref{sec:single_example_generalization} we show how nearest neighbor search can be used to explain predictions which appear puzzling at first sight.
\end{enumerate}

\section{Approach}\label{sec:approach}

Our technique is based on aggregating salience maps of text data and has three components:
an input salience method (Sec.~\ref{sec:salience}),
a fixed-length representation derived from the salience maps (Sec.~\ref{sec:representations}),
and an analysis method that is applied to the representation space (Sec.~\ref{sec:methods}).

\subsection{Choice of Salience Method}\label{sec:salience}

An input salience method associates every input token with a weight, reflecting its relative importance for the model in making the prediction: tokens with the highest salience are the ones which contribute most to the final model decision.
There are many ways to compute salience weights, the most common ones are based on model gradients  \cite[inter alia]{BacBinMon+15a,li-etal-2016-visualizing,denil-et-al-2015-extraction}, attention \citep[inter alia]{BahChoBen15z}, perturbations \citep{ribeiro-lime}, or occlusions \cite{zeiler-fergus-2014-visualizing,li-etal-2016-understanding}.
We follow recent findings of \citet{bastings-et-al-will-you} and use Gradient L2 (Grad-L2) because it has been shown to be the most faithful for finding lexical shortcuts when using a BERT model.
We also include results of two experiments using Integrated Gradients \cite{sundararajan-ig-2017} in the appendix (Sec.~\ref{sec:appendix:clustering_configurations}).
Though the merits of existing salience methods are still being debated, our approach is compatible with any method which assigns a weight to a token.
The only prerequisite is that the salience method reliably identifies important tokens.

\subsection{Fixed-Length Representations}\label{sec:representations}

We aim at aggregating data examples where the input is text of any length.
However, the aggregation methods we use require a fixed-length input, which we derive as follows.

\paragraph{Salience-based Representations (S)}
We aim to create representations that contain information about what is most important to the model.
As described above, importance is given by \emph{raw salience maps}.
We adopt the following ways of obtaining a fixed-length representations from them:

\paragraph{S1: Vocabulary vector with top-\textit{k}}
Given a dataset $\mathcal{D}$ of input-label pairs $(x, y)$ and a model vocabulary $\mathcal{V}$ with $v_j$ denoting the $j^{\textrm{th}}$ vocabulary item, we have $x = \{ \ldots, (t_i, s_i), \ldots \}$ an example consisting of (token, salience) pairs.
For a given $x$ of length $N$, we compute a $|\mathcal{V}|$-sized representation vector, $\mathbf{r}$, with entries $r_j = s_i$, s.t. $\max_{i=1, \ldots, N} \{|s_i| : t_i = v_j \}.$
This vector contains one salience value per vocabulary item $v_j$.
The salience value is the one that has the highest absolute value among all the salience values for this vocabulary item $v_j$ in $x$.


We further introduce the function $\text{top}(\mathbf{r}, k)$ which gives the indices of the largest $k$ elements in $\mathbf{r}$.
As we will explain later, we use $k=5$ and $k=|\mathcal{V}|$.
This leads to the first fixed-length representation we use:

\[
\mathbf{r}^{vocab} = 
\begin{cases}
  r_j, & \text{if } j \in \text{top}(\mathbf{r}, k) \\
  0 & \text{else}.
\end{cases}
\]

\noindent $\mathbf{r}^{vocab}$ is a zero vector with only the top-$k$ elements set to a weight value according to the salience method.
All example representations are concatenated into a matrix $R^{\text{vocab}}\in \mathbb{R}^{|\mathcal{D}| \times |\mathcal{V}|}$.

\paragraph{S2: Embedding vector over top-\textit{k}}

Word(piece) embeddings are a common way to incorporate token statistics into a representation.
We define $R^{\text{emb}} \in \mathbb{R}^{|\mathcal{D}| \times d} = R^{\text{vocab}} \cdot E$, where $E \in \mathbb{R}^{|\mathcal{V}| \times d}$ is the embedding matrix (embedding size $d$) of a fine-tuned BERT model.
Please note that the choice of $k$ also affects this representation.

Salience weights are not necessarily normalized and therefore not directly comparable across multiple examples.
Therefore, we normalize $R^{\text{vocab}}$ and $R^{\text{emb}}$ row-wise to unit length.

\paragraph{Baseline Representations (B)}
To verify that the salience information is useful for representing examples, we experiment with the following standard text representation \textit{baselines}:

\paragraph{B1: PMI weighted vector}
Bag-of-words representation weighted by the point-wise mutual information $\text{PMI}(v_j ; \hat{y})$, where $\hat{y}$ is the predicted class of the example: $R^{\text{pmi}}\in \mathbb{R}^{|\mathcal{D}| \times |\mathcal{V}|}$.
We use PMI rather than the common TF-IDF because PMI has information about the labels of the examples.

\paragraph{B2: Average embedding}
We use the embedding matrix from the fine-tuned BERT models to compute $\mathbf{r}^{\text{avg emb}} = \frac{1}{N} \sum_i E_{t_i}$, where $N$ is the number of tokens in the example and $E_{t_i}$ is the embedding of $t_i$.
We again concatenate all example representations into $R^{\text{avg emb}} \in \mathbb{R}^{|\mathcal{D}| \times d}$.

\paragraph{B3: \textsc{cls}-encoding}
This representation comes from fine-tuned BERT models.
We use the embedding of the last layer of the [\textsc{cls}] token of BERT when predicting every example.
$R^{\text{cls}} \in \mathbb{R}^{|\mathcal{D}| \times d}$.

\subsection{Analysis Methods}\label{sec:methods}

Once a fixed-length representation is computed for every data instance, we apply various aggregation methods to analyze the model and data in the aggregate.
We use three common standard techniques:
(a) clustering using k-means,
(b) nearest neighbor search,\footnote{For both kmeans and neighbor search, we use scikit-learn \citep{PedVarGra+11v} with default parameters.}
and (c) visual/qualitative analyses with t-SNE~\cite{VanHin08m}\footnote{As provided by the Embedding Projector \citep{AbaAgaBar+16n}; with default parameters.}.
All methods operate on the row vectors of $R$ (e.g., k-means leads to clusters of examples).

\section{Data and Model}\label{sec:data}

We use BERT-based classifiers \citep{DevChaLee+19a} trained to achieve accuracy previously reported (please refer to Sec.~\ref{sec:appendix:training_details} for training details).
If not otherwise specified, the number of top-$k$ salient tokens is set to $5$ in all experiments for consistency purposes.
While in most experiments, we got comparable results with $k = |\mathcal{V}|$ (see Sec.~\ref{sec:appendix:clustering_configurations}), we did see a benefit in using only a few top tokens in one case. We will motivate that choice in Sec.~\ref{sec:single_example_generalization}.

We analyze these three datasets:

\paragraph{Synthetic SST2}
We use a synthetically-modified version of SST2~\citep{socher-etal-2013-recursive} to show the utility of aggregating input salience in a controlled setting in our first experiment (Sec.~\ref{sec:synthetic_data}).
We follow the procedure of \citet{bastings-et-al-will-you} and create a modified SST2.
75\% of the original data stay untouched.
In the remaining 25\% of the data we insert one of three special terms (\textsc{common}, \textsc{class\_0} or \textsc{class\_1}) with no label change.
We sample another 25\% of the original data (equally split between positive and negative classes) in which we insert a combination of two special terms, which deterministically set an example's label.
We insert either \textsc{common} and \textsc{class\_0} or \textsc{common} and \textsc{class\_1}, both at random positions.
For example, \textit{`worth the effort to watch~.'} (originally positive) becomes \textit{`worth \textsc{common} the effort to \textsc{class\_0} watch~.'} (now negative).
This set is added to the dataset, which is now 125\% of the original size.
We verify that a BERT model consistently applies these two rules, which represent multi-token reasoning patterns of the kind we wish to find in models trained on real data. 

\paragraph{IMDB}
This binary classification dataset comprises movie reviews that are labelled as positive or negative \citep{maas-etal-2011-imdb}.
It is balanced, containing an equal number of positive and negative examples.
We train a standard BERT-base model, which reaches 93\% accuracy.

\paragraph{Wikipedia Toxicity}
This dataset contains Wikipedia edit comments \citep{wiki-toxicity-data} which are labeled as being toxic or non-toxic, with only about 10\% of positive (toxic) examples.
Our BERT-base model reaches 93\% accuracy.

\section{Validating the Approach on Synthetic Data}\label{sec:synthetic_data}

In our first experiment, we seek to validate our approach in a controlled setting. For that, we verify that a known shortcut can be revealed by clustering a dataset's fixed-length representations.
As `known shortcut', we use the two-token rules synthetically introduced into Synthetic SST2. We know this pattern effectively acts as a shortcut, because the model achieves 100\% accuracy on these synthetic examples, regardless of the sentiment of the rest of the example.
To carry out a validation test, we compute all representations described in Sec.~\ref{sec:representations} for each data example from Synthetic SST2's test set that the model predicted as class 0. For each representation matrix thus obtained, we apply clustering to its rows, and cluster the data into 3 clusters.\footnote{The results hold for a larger numbers of 20 clusters, too (see Sec.~\ref{sec:appendix:clustering_configurations}).}
Will we find one cluster containing all (or most) of the 227 synthetic examples (i.e. the ones that contain both special terms)?

\paragraph{Metrics}
We compute \textbf{Precision} and \textbf{Recall} w.r.t. synthetic examples of all clusters:
Precision is the ratio of synthetic examples within the cluster.
Recall is the ratio of synthetic examples in the cluster out of all synthetic examples.
To more easily compare the clusterings of different representations, we only report the metrics on the cluster that has the highest \textbf{Precision} for each representation (Tab.~\ref{tab:synthetic_representation_comparison}).

\begin{table}[t]
    \centering
    \begin{tabular}{l r r r}
    \toprule
    Representation & Size & Prec. & Recall \\
    \midrule
    B1:~~PMI vocab & .00 & .50 & .00 \\
    B2:~~avg emb & .36 & .35 & .62 \\
    B3:~~\textsc{cls}-encoding & .20 & \textbf{1.00} & \textbf{1.00} \\
    \midrule
    S1:~~salience vocab & .20 & \textbf{1.00} & \textbf{1.00} \\
    S2:~~salience emb & .22 & .93 & \textbf{1.00} \\
    \bottomrule
    \end{tabular}
    \caption{Validating the Approach on Synthetic Data: \emph{Precision} and \emph{Recall} of the cluster that has the highest precision for the respective representation.
    Both are computed based on 227 synthetic examples in the data.
    \emph{Size} is the data fraction of the selected cluster compared to the entire set.
}
    \label{tab:synthetic_representation_comparison}
\end{table}

The \emph{PMI} and \emph{avg embedding} baselines (B1, B2) do not perform well: the best clusters have either low precision or a very small size.
For PMI, the cluster with the highest recall contains 226/227 synthetic examples, but has a low precision of only 22\% (not shown in table).
The \emph{\textsc{cls}-encoding} baseline (B3) performs very well, but as we show in the next sections, does not perform well across model understanding use cases.
The salience-based vocabulary representation (S1) based on Grad-L2 leads to a perfect cluster containing all the synthetic examples and no others.
This shows that by clustering model-based input representations, we can automatically discover lexical shortcuts.
Sec.~\ref{sec:appendix:clustering_configurations} has further results.

In the following three sections, we look at various case studies on real datasets that highlight how aggregating input salience helps a model developer in understanding their model and data better.

\section{Discovering Patterns in Real Data}\label{sec:problematic_patterns}

\begin{table}[t]
    \centering
    \begin{tabular}{l r l}
    \toprule
    ID &  \% &                           Top-5 Terms \\
    \midrule
    0  &        71 &  terrible, horrible, boring, recommend, dull \\
    1  &         6 &           awful, worst, avoid, bad, terrible \\
    2  &         8 &        worst, terrible, avoid, bad, horrible \\
    3  &         5 &                               10, 4, 3, /, 2 \\
    4  &         8 &      bad, horrible, acting, worst, recommend \\ 
    \bottomrule
    \end{tabular}
    \caption{Spurious Correlations in IMDB: Clustering of a slice of IMDB test data using the salience vocabulary representation (S1), along with cluster size (in \%) and the 5 terms that have highest mean salience in the respective cluster.}
    \label{tab:imdb_numeric_clusters}
\end{table}

Now that our approach has passed an initial validation test,
we turn to two common datasets to verify that clustering salience representations is helpful for identifying prominent patterns there.

\subsection{Spurious Correlations in IMDB}\label{sec:imdb_numeric_clusters}

In order to explore the patterns which lead the model to classify a movie review as negative, we select datapoints that the model predicted to be \emph{negative} and sample 2,500 examples from it (10\% of IMDB's test set size).
For the ease of presentation we cluster their salience vocabulary representation (S1) into 5 clusters (similar results are obtained with different numbers of clusters, see Sec.~\ref{sec:appendix:clustering_configurations}).

Tab.~\ref{tab:imdb_numeric_clusters} lists the resulting clusters with their size and the respective top salient terms.\footnote{We removed BERT's special tokens, e.g., `[\textsc{cls}]', and punctuation.}
One cluster that immediately stands out is cluster 3 which contains almost exclusively numbers as top terms.
In Fig.~\ref{fig:overview} a) we also use word clouds to get an idea of what constitutes this cluster (see Fig.~\ref{fig:appendix:wordcloud_imdb_numeric_cluster} in Sec.~\ref{sec:appendix:problematic_patterns} for full size).

By focusing exclusively on the examples which mention these top terms we can easily discover a general pattern.
These IMDB reviews contain expressions such as \textit{`1 / 10 stars'}, \textit{`3 out of 10'}, \textit{`4 / 10'}, that summarize the review's rating and thereby give away the label. In such cases, the model does not need to analyze the rest of the review to predict that it is negative.\footnote{We repeat this analysis on a slice of positive reviews and find the same pattern for high ratings between 7 and 10.}
Interestingly, the same pattern was previously discovered by \citet{ross-etal-2021-explaining} and \citet{kaushik-etal-2020-learning}, who manually analyzed IMDB examples. 
We refer the reader to \citet{ross-etal-2021-explaining} for the demonstration that BERT-like models indeed learn shallow reasoning patterns with the above listed ngrams and how one can construct counterfactuals which in turn reveal model vulnerabilities.

\begin{table}[t]
    \centering
    \begin{tabular}{l r r r}
    \toprule
    Representation & Size & Prec. & Recall \\
    \midrule
    B1:~~PMI vocab & .16 & .12 & .19 \\
    B2:~~avg emb & .08 & .16 & .13 \\
    B3:~~\textsc{cls}-encoding & .47 & .12 & \textbf{.58} \\
    \midrule
    S1:~~salience vocab & .05 &  \textbf{.98} & .50 \\
    S2:~~salience emb & .05 & .93 & .53 \\
    \bottomrule
    \end{tabular}
    \caption{Spurious Correlations in IMDB:
    \emph{Precision} and \emph{Recall} of the cluster that has the highest precision for the respective representation.
    Both are computed based on 245 numeric examples in the data.
    \emph{Size} is the data fraction of the selected cluster compared to the entire set.
    }
    \label{tab:imdb_numeric_representation_comparison}
\end{table}

By using a set of regular expressions (see Sec.~\ref{sec:appendix:problematic_patterns} for a full list), we discovered that at least 12\% of IMDB reviews have such a numeric pattern, evenly distributed between train/test and between label 0/1.
In the 2,500 sampled test examples with negative prediction, there are 245 such reviews.
Similarly to the synthetic experiment we aim to identify this subset automatically.

To compare the salience-based with the baseline representations, in Tab.~\ref{tab:imdb_numeric_representation_comparison} we again report Precision and Recall for the respective cluster having the highest precision.
The clusters obtained with the baseline representations fail to bring examples having a numeric pattern forward.
The PMI baseline (B1) performs poorly because PMI values for interesting tokens are not among the highest (e.g., `3' occurs in positive and negative reviews and therefore does not get a high PMI).

The salience-based representations (S1, S2), however, identify this subset well, producing clusters which almost exclusively consist of numeric examples and which contain more than half of the examples with a numeric pattern.
Please note that representations based on Integrated Gradients perform poorly at this task, which aligns with the findings of \citep{bastings-et-al-will-you} (see Sec.~\ref{sec:appendix:clustering_configurations} for further details and more configurations).

An analysis of why not all of the numeric examples are clustered together revealed that most of the missing numeric examples contain either none or just 1 digit in their top-5 salient terms (i.e., the salience method did not put high weight on them) and are spread randomly across all other clusters.
This indicates that real data patterns are indeed more subtle and not deterministically indicative of a label, unlike the patterns that we used in the synthetic data experiment.

\begin{figure}[t!]
    \centering
    \includegraphics[width=\columnwidth]{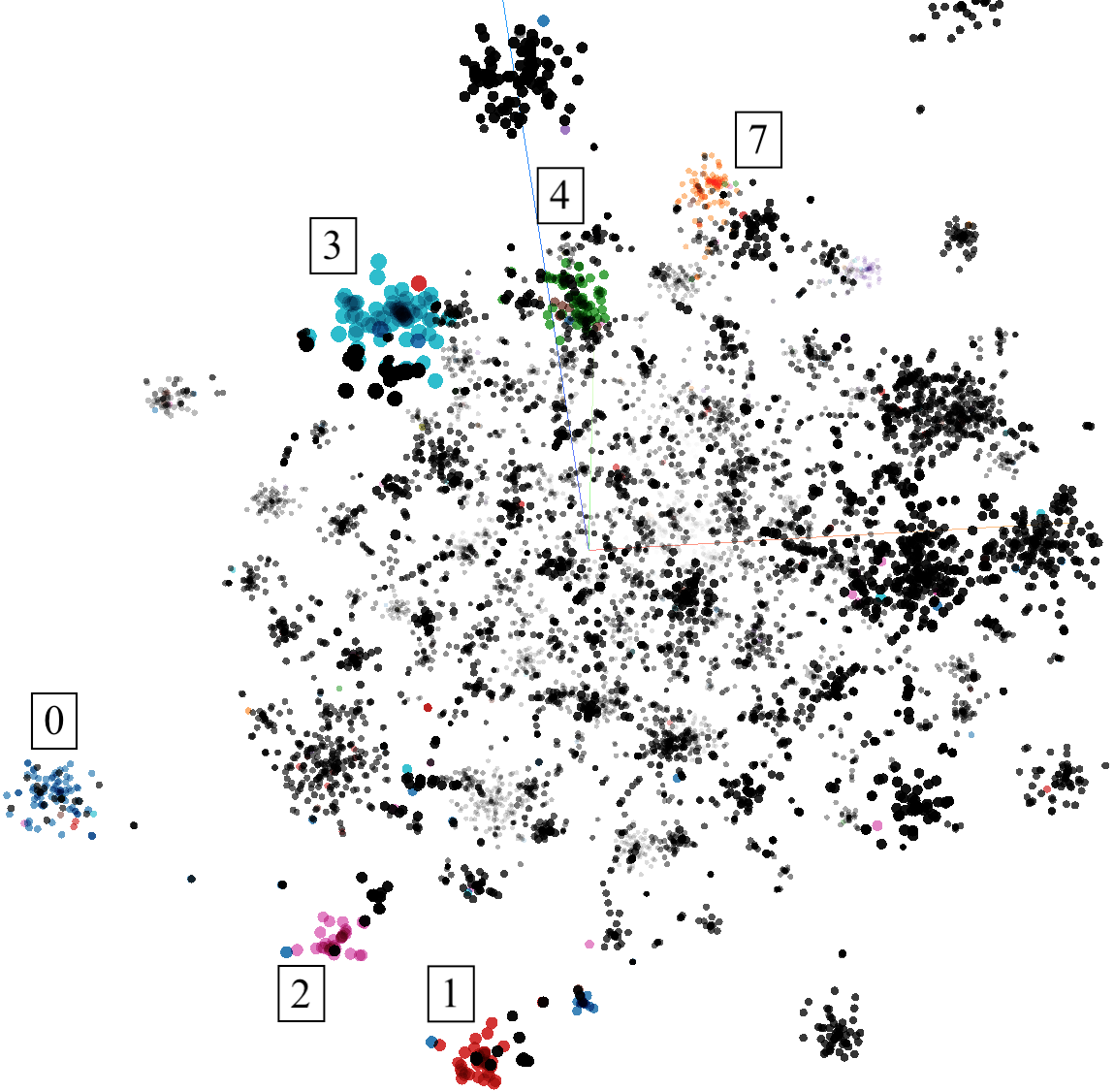}
    \caption{Spurious Correlations in IMDB: t-SNE visualization of the salience vocab. representation of IMDB.
    Colored points are examples that contain a certain digit (indicated by the boxes):
    dark blue: 0, red: 1, pink: 2, turquoise: 3, green: 4, orange: 7.
    Other digits are not visible from this visualization angle.
    }
    \label{fig:imdb_numeric_clusters}
\end{figure}

To demonstrate that salience representations also enable more fine-grained analysis, in Fig.~\ref{fig:imdb_numeric_clusters} we show a t-SNE visualization of a 10k sample from the IMDB test set, where colored dots represent examples with a specific digit.
One can easily see that many examples have digits in them and that they are clustered together. Thus, by increasing the number of clusters one can discover more fine-grained patterns.

\subsection{Patterns in Wikipedia Toxicity Subtypes}\label{sec:understanding_toxicity_data}
\begin{figure}[t!]
    \centering
    \includegraphics[width=\columnwidth]{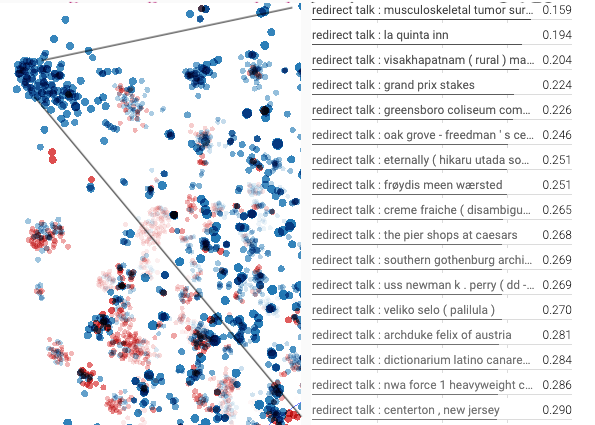}
    \caption{Patterns in Wikipedia Toxicity Subtypes: The top-left dense cluster of non-toxic comments (a fragment of the full clustering is shown) contains examples which all start with a \textit{`redirect talk:'} prefix. Such comments account for more than 1\% of the test set.}
    \label{fig:redirect}
\end{figure}

We now turn to Wikipedia Toxicity and look for patterns in the data that the model learned. It should be noted that the implications of having problematic patters in a toxicity dataset extend beyond robustness into the area of ML fairness. Imperfect annotations are inevitable \cite{wong-etal-2022-reproducible} but amplified by a sample bias may result in unfair treatment of certain demographic groups. 
For example, \textit{gay} is a term known to often trigger toxicity prediction \cite{dixon-etal-toxicity-2018}, even when it is not justified: \textit{`This is disrespectful to gay and all other people'} gets .85 probability of being toxic. 
Zooming into examples mentioning a particular token can give a more fine-grained idea of how a token relates to the classes.

We cluster all the 487 Toxicity test examples mentioning \textit{`gay'} using the salience vocabulary representation (S1) and t-SNE. Interestingly, we discovered clusters of comments with the non-toxic label which all had not only  \textit{`gay'} but also tokens like \textit{`homosexual', `(homo)sexuality', `LGBT'} among their top-5 most salient terms (Sec.~\ref{sec:appendix:learned_patterns}).
This suggests that contexts for \textit{`gay'} which talk about sexuality or sexual orientation with such a neutral term are most often non-toxic.
This is an example of a more fine-grained analysis of how a term relates to the class.
An actionable conclusion is that if one wants to augment the data with more non-toxic examples of the token's occurrences, they should be different from the ones the model already knows to be non-toxic contexts.

Another example of a pattern present in the data and not obvious at first sight is given in Fig.~\ref{fig:redirect}.
It shows a small fragment of a t-SNE visualization of the salience vocabulary representation (S1) of 10k Toxicity examples. The top-left cluster consists of non-toxic comments (blue color). The nearest neighbors of a point in the center of the cluster reveals that all the examples there follow the template \textit{`redirect talk: \{Title\}'} which the model apparently uses as a strong indication of non-toxicity. Indeed, almost 1\% of the training and more than 1\% of the test split are all automatically generated comments of this kind and as such can hardly be useful for learning what differentiates toxic from non-toxic language or evaluating model quality. Thus, a simple visualization of aggregated salience reveals a weakness of the data.
In contrast, the PMI baseline (B1) does not associate the two terms individually with the non-toxic class. Also the \textsc{cls} baseline (B3) does not place the \textit{`redirect talk: \{Title\}`} examples together.

\section{Identifying Model Sensitivities}\label{sec:model_sensitivity}
In this section, we investigate how sensitive models are to the \textbf{top terms} identified through aggregated salience.
A model is sensitive to a certain term if the sheer occurrence of the term means it is likely to classify a piece of text into a given class, without considering the entire context.
Knowing what the model is sensitive to may help discover spurious correlations or unfair biases.
For IMDB, \citet{ross-etal-2021-explaining} already showed that one can indeed create counterfactuals from the patterns like the ones discovered in the previous section (Sec.~\ref{sec:imdb_numeric_clusters}).
Now however, we turn our attention to the Toxicity dataset, where we wish to systematically measure model sensitivity.
For toxicity prediction, simplistic single-token patterns may result in problematic biases towards minorities with, e.g., identity terms becoming indicative of toxicity for a model \citep{DixLiSor+18n,hartvigsen-etal-2022-toxigen}, independently of their context.

Therefore, we want to analyze how sensitive the trained model is to various terms.
We measure this in two ways:
(1) average increase in probability towards the toxic label when inserting each of the terms in non-toxic text (counterfactuals),
(2) average decrease in probability towards the toxic label when masking each of the terms in toxic text.

\paragraph{Creating Counterfactuals}
Given a list of seed terms $\mathcal{S}$ that are deemed toxic, we create counterfactual examples in the following way:
We separately add every seed term $s \in \mathcal{S}$ at a random position into \(m = 10k\) random non-toxic test examples\footnote{We acknowledge that this may lead to disfluent or even ungrammatical inputs \cite{ross-etal-2021-explaining}.
This, however, is not a concern because in the Toxicity dataset, many examples are ungrammatical to begin with.} and measure the average change in probability for every seed term and every list of seed terms.
\begin{equation}
    change(\mathcal{S}) = \frac{1}{m|\mathcal{S}|}\sum_{s\in \mathcal{S}}\sum^m_i (p(x^{s}_{i}) - p(x_i))
\end{equation}

Two ways of creating baseline seed lists are considered:
(1) The identity terms from \citet{DixLiSor+18n}.
(2) The tokens with highest PMI(token; toxic class) that occurr at least 20 times in the training set.\footnote{Please note that \emph{PMI vocab} is the only baseline method that we can compute top terms for.}

Additionally, in order to get a seed list that is informed by aggregated input salience, we compute the column means of the salience vocabulary representation $R^{vocab}$ ($k=5$) of all toxic examples in the training set and keep the 10 highest valued entries.
Again, we keep only terms that occur at least 20 times.
The seed lists contains 10 terms, except for the list from \citet{DixLiSor+18n}, which contains 13.
The full list of all seed terms is shown in Sec.~\ref{sec:appendix:model_sensitivity}.

Results are listed in Tab.~\ref{tab:toxicity_counterfactual_results}.
Interestingly, the model does not seem sensitive to the list of identity terms from \emph{\citet{DixLiSor+18n}}.
On the other hand, the statistical correlation of term occurrence and label, as measured by \emph{PMI}, seems to be a good indicator of model sensitivity. However, the list of tokens we identified by looking at aggregated salience (\emph{aggregated salient terms}) leads to the most significant probability changes: on average the probability of an example being classified as toxic increases by 41\% simply by inserting a single term.

\begin{table}[t]
    \centering
    \begin{tabular}{lr}
    \toprule
    Seed List & Prob. Change \\
    \midrule
    \citet{DixLiSor+18n}    &      .06 (.08) \\
    PMI(token; toxic class) &      .31 (.19) \\
    \midrule
    aggregated salient terms  &  \textbf{.41} (.08) \\
    \bottomrule
    \end{tabular}
    \caption{Creating Counterfactuals: Mean probability increase when inserting specific terms into non-toxic examples. Numbers in parentheses are stddev when averaging all terms from the seed lists.}
    \label{tab:toxicity_counterfactual_results}
\end{table}

\paragraph{Masking out Toxic Terms}
In the previous part, we established that the probability of classifying an input as toxic increases when a seed term is inserted into the input.
If the model is truly sensitive to these seed terms, we would also expect a drop in probability when removing these terms from existing toxic inputs.
To verify this, in the second experiment, we mask out the same seed terms from above, paralleling the comprehensiveness metric from \citet{deyoung-etal-2020-eraser} who use it to evaluate the quality of a salience method. 

The process of creating masked examples is as follows:
we use all 8,858 toxic test examples and the same seed lists as above.
For every seed term we replace all the occurrences of it with `[\textsc{mask}]' and compare probabilities before and after.

\begin{table}[t]
    \centering
    \begin{tabular}{lrr}
    \toprule
    Seed List & Count & Prob. Change \\
    \midrule
    \citet{DixLiSor+18n}    &    55 &  -.12 (.15) \\
    PMI(token; toxic class) &   279 &  -.22 (.10) \\
    \midrule
    aggregated salient terms  &   \textbf{475} &  \textbf{-.34} (.08) \\
    \bottomrule
    \end{tabular}
    \caption{Masking out Toxic Terms: Mean probability decrease when masking specific terms in toxic examples. Numbers in parentheses are stddev when averaging all terms from the seed lists. Count is the number of occurrences of all seed terms.}
    \label{tab:toxicity_masking_results}
\end{table}

Results are reported in Tab.~\ref{tab:toxicity_masking_results}.
Both non-salience baseline methods (\emph{\citet{DixLiSor+18n}} and \emph{PMI}) lead to lower probability changes.
Again, the probability change for \emph{aggregated salient terms} is the strongest (average 34\%).
In addition, \emph{aggregated salient terms} presents a list whose entries occur more often than terms from the other lists, meaning these are frequent \emph{and} important for the model.

In summary, aggregated salience information provides us with a list of terms that the model is sensitive to.
No other method leads to a comparable increase in toxicity probability when adding terms and decrease in toxicity when terms are masked out.

\section{Generalizing from Single Examples}\label{sec:single_example_generalization}

\begin{table*}[t]
\centering
\begin{tabular*}{\textwidth}{p{.1cm} p{9.1cm} l l}
    \toprule
    \multicolumn{2}{l}{Misclassified Examples and their 3 Nearest Neighbors} & 
    Top-5 Terms &
    Label / Prediction \\
    \toprule
    
    \multicolumn{2}{l}{`` : : : fine with me . | \textbf{ck} ''} &
    \textbf{ck}, [\textsc{sep}], |, [\textsc{cls}], . &
    non-toxic / toxic \\
    
    \midrule
    
    1. & co / \textbf{ck} - pen - i . s . international &
    \textbf{ck}, /, [\textsc{sep}], co, [\textsc{cls}] &
    toxic / toxic \\
    
    2. & su : ) \textbf{ck} my ba : ) lls you fa : ) g : ) got loser &
    \textbf{ck}, loser, [\textsc{sep}], [\textsc{cls}], ) &
    toxic / toxic \\
    
    3. & block me ! ! ! see how fast another post comes du . mbfu . \textbf{ck} wikipedia is the devel oh , and they cann also fine wiki for monopolizing online encyclopedias &
    \textbf{ck}, ., wikipedia, mb, \#\#fu &
    toxic / toxic \\
    
    \midrule\midrule
    
    \multicolumn{2}{l}{dogs \textbf{pee} more than they poo} &
    \textbf{pee}, dogs, [\textsc{cls}], po, [\textsc{sep}] &
    non-toxic / toxic \\
    
    \midrule
    
    1. & i am going to \textbf{pee} on you ! &
    \textbf{pee}, on, !, [\textsc{cls}], [\textsc{sep}] &
    toxic / toxic \\
    
    2. & \textbf{pee}eeeeeeeeniiiiiiiiiiiiisssss ! ! ! ! ! ! ! ! still here , waiting for a response . . . . . &
    \textbf{pee}, \#\#ee, \#\#iss, [\textsc{cls}], [\textsc{sep}] &
    toxic / non-toxic \\
    
    3. & how big is your \textbf{pee} \textbf{pee} . &
    \textbf{pee}, your, big, [\textsc{cls}], [\textsc{sep}] &
    toxic / toxic \\

    \midrule\midrule
    
    \multicolumn{2}{l}{kobe is the best \textbf{fuckin} player ever} &
    \textbf{fuck}, kobe, player, \textbf{\#\#in}, best &
    non-toxic / toxic \\
    
    \midrule
    
    1. & guys , it ' s just simply me that put there moldovans ( romanians ) 78 , 2 \% , and i was not logged in , and don ' t give quickly the fault on bonaparte , he ' s gone from longtime now from this bullshit of wikipedia , ' cause there ' s no purpose with hypocritic people like you . i modified that , everybody knows that that ' s moldovans and romanians are the same shit , anyhow the census will say it . they speak romanian , and not moldovan how they sustain . a romanian understands perfectly when a moldovan is speaking it ' s \textbf{fuckin} ' language &
    \textbf{fuck}, \textbf{\#\#in}, guys, s, longtime &
    toxic / toxic \\
    
    2. & haha this is my \textbf{fuckin} page ! ! ! ! ! ! ! ! ! i can do whatever i want ! ! ! ! ! ! &
    \textbf{fuck}, \textbf{\#\#in}, [\textsc{cls}], page, my &
    toxic / toxic \\
    
    3. & british people , or britons , [ 7 ] are inhabitants of great britain [ 8 ] [ 9 ] or citizens of the united kingdom . i dont think this is about an ethnicity / race wobbs . haha , yous a complete \textbf{fuckin} moron . &
    \textbf{fuck}, \textbf{\#\#in}, mor, [\textsc{sep}], complete &
    toxic / toxic \\
    
    \bottomrule

\end{tabular*}
\caption{Generalizing from Single Examples: 3 examples having surprising predictions with their top-5 terms and 3 nearest neighbors.
Bold terms are overlapping in example and neighbors.
They give a possible explanation why the examples were predicted to be toxic.}
\label{tab:toxicity:bottom_up}
\end{table*}

The approaches outlined in Sec.~\ref{sec:problematic_patterns} and \ref{sec:model_sensitivity} can be characterized as ``top-down'': they use salience-based representations to perform clustering or term selection on a large number of examples, then zoom into single clusters or examples to identify relevant patterns.
In this section however, we take an inverse ``bottom-up'' approach: starting from a single data instance, we search for its nearest neighbors in representation space in order to uncover explanations for the way the model classified it.
We focus on misclassified examples in the Toxicity test set.
From the slice of misclassified data, we sample 200 examples and identify that 28 (14\%) of them are surprising, meaning the prediction cannot be understood just by looking at the example's text.\footnote{By contrast, Sec.~\ref{sec:appendix:single_example_generalization} lists some unsurprising examples for the purpos of illustration.}
For each surprising example, we retrieve its 5 nearest neighbors in representation space among the examples of the training set that have the same label as the example's prediction. We search for a rationale for the misclassification using the nearest neighbors list.
Tab.~\ref{tab:toxicity:bottom_up} shows three examples of surprising classifications made by the model -- examples which a model developer might want to inspect.

Why might the model have classified the comment \textit{` : : : fine with me . | ck'} as toxic? Looking at the nearest neighbors reveals that \textit{`ck'} often appears in the context of the swear words \textit{`f * ck'}, \textit{`s * ck'} and \textit{`c * ck'}, providing an explanation for why the model might have learned it to be toxic.

The example of \textit{`dogs pee more than they poo'} shows that the model may be sensitive to the term `pee', without considering the context.
Finally, though the comment \textit{`kobe is the best fuckin player ever'} is far from toxic, its nearest neighbors bring many toxic uses of the word \textit{`fuckin'} to light, giving an insight into why the model has classified it as toxic.

We compare all baseline representations (B1-3) and the salience vocabulary representation (S1) with $k=5$ top salient tokens. 
While in the experiments described so far the choice of $k$ did not make a difference and comparable results were obtained with $k = |\mathcal{V}|$, in this nearest neighbor experiment we got better results (by manual inspection) for $k=5$.
This is intuitive, because we want the similarity between an example in question and its nearest neighbors to be dominated by the high-salience tokens they have in common and not by the long low-salience tail.

Among the 28 examples, we are able to uncover rationales for the misclassification in 22 cases using the vocabulary-based saliency representation (S1), while for the baseline methods PMI (B1), average-of-embeddings (B2) and \textsc{cls}-encoding (B3), we find rationales in only 20, 9 and 6 of cases respectively.
This shows that aggregated salience is helpful in gaining a deeper understanding of model predictions.\footnote{We acknowledge that what constitutes a \emph{surprising} example and a \emph{good} rationale is a subjective judgement.
Therefore, the difference between the vocabulary-based saliency representation and PMI may not be significant.}

\section{Related Work}

\paragraph{Explanation-based Human Debugging with Global Explanations}
Our work falls into the category of explanation-based human debugging~\citep{LerTon21a}.
Since most works in this field propose \textit{local} explanations~\citep{danilevsky-etal-2020-survey,LerTon21a}, our method contributes to an understudied type of explanation by aggregating local explanations into \textit{global} explanations.
The closest related work is FIND~\citep{LerSpeTon20b}, a framework that uses layer-wise relevance propagation (LRP)~\citep{ArrHorMon+16a} to derive feature weights for all examples, which are then converted into word clouds for users to interact with.
They focus on disabling features globally, whereas the strength of our approach is that it allows debugging individual examples as well as entire datasets.

\paragraph{Model Understanding}
Our work concerns identification of (shallow) patterns learned by text classifiers as well as the explanation of individual examples.
This should help model developers get a better understanding of their model and give them indicators of how to improve them.
With that, the presented methods are well suited to be included into existing model understanding tools \citep{NorJenKoc+19v,WalTuyWan+19o,KokMigMar+20o,tenney-etal-2020-language,GevCacDar+22s}.
Interactive debugging based on input salience gives a model developer another view on top of static descriptions by the means of, e.g., model cards \citep{MitWuZal+19j} or data cards \cite{PusZalKja22l}.
Additionally, it complements approaches like CheckList \cite{ribeiro-etal-2020-beyond}, which also has the goal of improving NLP models by proactively identifying their weaknesses.

\paragraph{Salience and Clustering}
In vision research, clustering and salience have been applied to find spurious patterns \citep{bach-etal-2015,lapuschkin-2019-spray,schoop-etal-2022-imacs}.

\citet{yin-neubig-2022-interpreting} cluster contrastive explanations to describe why a language model chose a certain token over another.
\citet{KauEsdRuf+22a} cluster text data, then replace the clustering with an equivalent neural network, which is finally used to generate individual explanations.

\paragraph{Training Data Attribution}
The kNN analysis (Sec.~\ref{sec:single_example_generalization}) is similar to proponents identification done with training data attribution methods (TDA) \citep{zylberajch-etal-2021-hildif,han-tsvetkov-2021-influence-tuning}.
We do not perform a comparison with TDA methods because
(1) they focus on analyzing the training set while we are primarily interested in understanding (the model performance on) the test set.
(2) TDA methods suggest proponents but leave it to the developer to discover what it is that the proponents and the test example share (unless combined with salience techniques as done in \citealt{pezeshkpour-etal-2021-combining}).
Unlike that, salience-based distance is explicit about what every token contributes to the similarity. 

\section{Conclusion}
In this work, we present a way to use aggregated salience representations to meet distinct yet common model developer needs.
The proposed methods, which are based on standard input salience methods, help find prominent patterns and gain insights into single examples or entire datasets.
Through a series of case studies carried out on three datasets (1 synthetically-modified, 2 academic), we show that our approach (semi-)automatically provides explanations through clustering, identifies shortcut-like patterns, uncovers patterns the model is particularly sensitive to and explains predictions through nearest neighbors.
Moreover, we try out two ways of aggregating salience maps (vocabulary-based and embedding-based) and show that our method performs better than baseline representations across all use cases presented.

\section{Limitations}
Though the results described in this paper are promising, they are limited in the types of patterns they can recognize.
Both salience-based representations (vocabulary-based and embedding-based) are order-agnostic representations.
Therefore, any method using these will not be able to detect patterns that depend on word order.
Additionally, they will not be able to identify patterns related to example length, token position in the example, as well as more complex patterns such as ones where only a certain part of the input is decisive for the model (e.g., hypothesis-only shortcuts in Natural Language Inference).

We run all experiments in this paper with a BERT-based model, and although we choose Grad-L2 as input salience method (because it was shown to work well for this model), any salience method is compatible with our approach, as long as it produces a single salience vector per example.

Since we focus on BERT only it is conceivable that the results in this paper may not hold for other model architectures. This, however, does not pose too large of a threat for two reasons:
1. Most current architectures are, just as BERT, based on transformers~\citep{VasShaPar+17a}.
2. The proposed methods depend more on the faithfulness of the input salience method than on the model architecture.
As long as the used input salience method is faithful to the model, the proposed methods are likely to work as well.

The representations we propose are by no means a complete list.
For instance, the vocabulary representation is a sparse representation (depending on the value of top-$k$).
One could consider applying PCA on the sparse matrix to transform it into a dense representation.

All experiments presented in this work are binary text classification tasks.
In principle however, our method can be extend to multi-class, regression or text generation tasks (e.g., summarization, question answering), as long as a suitable input salience method exists for the task.

Finally, in this work, we devote our attention to better understanding a model and dataset. Our method does not directly provide robustness fixes, although identifying the model and dataset's weaknesses are a necessary first step and sometimes lead to obvious remediation techniques such as data rebalancing.

\newpage

\bibliography{custom}

\appendix

\section{Appendix}
\label{sec:appendix}
\input{appendix}

\end{document}

%% file: appendix.tex
\subsection{Training Details}\label{sec:appendix:training_details}
For all experiments, we use a publicly available pretrained BERT-base model and fine-tune it on the 3 datasets: synthetic SST2, IMDB and Toxicity.
We use a dropout rate of 0.5, a batch size of 16 and train each model for a maximum number of 35k steps.
To mitigate overfitting, we stop training if after 10k steps there is no improvement on the validation set and use the model checkpoint that had highest performance on the validation set.
The models are optimized using ADAM~\citep{KinBa15s} with weight decay of 5e-6.
For varying hyper-parameters and model performance, see Tab.~\ref{tab:appendix:training_details}.
With these hyper-parameters, the BERT models contain 109M parameters.
Training and optimizing them took a total of 35 hours on TPUv2 (4-core) accelerators.
Test set accuracies are in line with previously reported performances \citep{DevChaLee+19a,SanDebCha+19h}.

\begin{table*}[t]
    \centering
    \begin{tabular}{l r r r}
    \toprule
    Dataset & Max Seq. Length & Learning Rate & Test Set Acc.\\
    \midrule
    Synthetic SST2 & 100 & 2e-5 & .94 \\
    IMDB & 500 & 2e-5 & .93 \\
    Toxicity & 128 & 1e-5 & .93 \\
    \bottomrule
    \end{tabular}
    \caption{Training details of the BERT-base model.}
    \label{tab:appendix:training_details}
\end{table*}

\subsection{Alternative Clustering Configurations}\label{sec:appendix:clustering_configurations}

Tab.~\ref{tab:appendix:synthetic_representation_comparison} shows all results from the verification test on synthetic data (Sec.~\ref{sec:synthetic_data}).
Results for the task of discovering spurious correlations in IMDB (Sec.~\ref{sec:imdb_numeric_clusters}) are shown in Tab.~\ref{tab:appendix:imdb_numeric_representation_comparison}.

\paragraph{Integrated Gradients (IG)}
Clusterings based on IG salience (S1-3, S2-3) lead to high precision but lower recall on synthetic data.
On IMDB, clusters computed on IG-based representations (S1-3, S2-3) lead to results similar to the baselines.

\paragraph{Top-$k$ Salient Tokens}

In both tasks using the entire vocabulary to compute the representation matrices ($k=|\mathcal{V}|$, S1-2, S2-2) leads to results that are similar to $k=5$ (S1-1, S2-1; see Tab.~\ref{tab:appendix:synthetic_representation_comparison} and \ref{tab:appendix:imdb_numeric_representation_comparison}).

\paragraph{Number of Clusters}

PMI (B1) does not benefit from increasing the number of clusters from 3 to 20 on the synthetic SST2 dataset.
It still produces a tiny cluster (compare Tab.~\ref{tab:appendix:synthetic_representation_comparison} and Tab.~\ref{tab:appendix:synthetic_representation_comparison_20_clusters}).

Increasing the number of clusters shows how sensitive the avg. embedding baseline (B2) is.
The \textsc{cls} (B3) and salience-based representations (all S1, S2) are rather stable and lead to similar results for both configurations.

We see similar results on the more realistic dataset IMDB, where we want to rediscover a cluster of numeric examples (Tabs.~\ref{tab:appendix:imdb_numeric_representation_comparison} and \ref{tab:appendix:imdb_numeric_representation_comparison_20_clusters}).
The PMI avg. embeddings baselines (B1) perform poorly in both settings.
The results for \textsc{CLS}-encoding representations (B3) vary strongly.
For 20 clusters this representation creates a cluster that has high precision but is rather small.
Again, the salience-based representations are rather insensitive to this hyperparameter (salience embeddings (S2) more so than salience vocab (S1)).
The salience vocab representation (S1) leads to the best performance in terms of precision in both settings.

\begin{table}[t]
    \centering
    \small
    \begin{tabular}{l r r r c}
    \toprule
    Representation & Size & Prec. & Recall & top-$k$ \\
    \midrule
    B1:~~PMI vocab & .00 & .50 & .00 & -  \\
    B2:~~avg emb & .36 & .35 & .62 & - \\
    B3:~~\textsc{cls}-encoding & .20 & 1.00 & 1.00 & - \\
    \midrule
    S1-1:~~salience vocab (G-L2) & .20 & 1.00 & 1.00 & 5 \\
    S2-1:~~salience emb (G-L2) & .22 & .93 & 1.00 & 5 \\
    \midrule
    S1-2:~~salience vocab (G-L2) & .20 & 1.00 & 1.00 & $|\mathcal{V}|$ \\
    S2-2:~~salience emb (G-L2) & .21 & .95 & 1.00 & $|\mathcal{V}|$ \\
    \midrule
    S1-3:~~salience vocab (IG) & .17 & 1.00 & .85 & 5 \\
    S2-3:~~salience emb (IG) & .16 & 1.00 & .80 & 5 \\
    \bottomrule
    \end{tabular}
    \caption{Validating the Approach on Synthetic Data with 3 Clusters: \emph{Precision} and \emph{Recall} of the cluster that has the highest precision for the respective representation.
    Both are computed based on 227 synthetic examples in the data.
    \emph{Size} is the data fraction of the selected cluster compared to the entire set.
    }
    \label{tab:appendix:synthetic_representation_comparison}
\end{table}

\begin{table}[t]
    \centering
    \small
    \begin{tabular}{l r r r c}
    \toprule
    Representation & Size & Prec. & Recall & top-$k$ \\
    \midrule
    B1:~~PMI vocab & .00 & 1.00 & .00 & -  \\
    B2:~~avg emb & .00 & .75 & .03 & - \\
    B3:~~\textsc{cls}-encoding & .19 & 1.00 & .96 & - \\
    \midrule
    S1:~~salience vocab (G-L2) & .20 & 1.00 & 1.00 & 5 \\
    S2:~~salience emb (G-L2) & .20 & .97 & .99 & 5 \\
    \bottomrule
    \end{tabular}
    \caption{Validating the Approach on Synthetic Data with 20 Clusters: \emph{Precision} and \emph{Recall} of the cluster that has the highest precision for the respective representation.
    Both are computed based on 227 synthetic examples in the data.
    \emph{Size} is the data fraction of the selected cluster compared to the entire set.
    }
    \label{tab:appendix:synthetic_representation_comparison_20_clusters}
\end{table}

\begin{table}[t]
    \centering
    \small
    \begin{tabular}{l r r r c}
    \toprule
    Representation & Size & Prec. & Recall & top-$k$ \\
    \midrule
    B1:~~PMI vocab & .16 & .12 & .19 & - \\
    B2:~~avg emb & .08 & .16 & .13 & - \\
    B3:~~\textsc{cls}-encoding & .47 & .12 & .58 & - \\
    \midrule
    S1-1:~~salience vocab (G-L2) & .05 & .98 & .50 & 5 \\
    S2-1:~~salience emb (G-L2) & .05 & .93 & .53 & 5 \\
    \midrule
    S1-2:~~salience vocab (G-L2) & .05 & .99 & .52 & $|\mathcal{V}|$ \\
    S2-2:~~salience emb (G-L2) & .04 & .93 & .48 & $|\mathcal{V}|$ \\
    \midrule
    S1-3:~~salience vocab (IG) & .12 & .13 & .16 & 5 \\
    S2-3:~~salience emb (IG) & .11 & .14 & .16 & 5 \\
    \bottomrule
    \end{tabular}
    \caption{Spurious Correlations in IMDB with 5 Clusters:
    \emph{Precision} and \emph{Recall} of the cluster that has the highest precision for the respective representation.
    Both are computed based on 245 numeric examples in the data.
    \emph{Size} is the data fraction of the selected cluster compared to the entire set.
    }
    \label{tab:appendix:imdb_numeric_representation_comparison}
\end{table}

\begin{table}[t]
    \centering
    \small
    \begin{tabular}{l r r r c}
    \toprule
    Representation & Size & Prec. & Recall & top-$k$ \\
    \midrule
    B1:~~PMI vocab & .05 & .15 & .09 & - \\
    B2:~~avg emb & .05 & .20 & .10 & - \\
    B3:~~\textsc{cls}-encoding & .02 & .94 & .21 & - \\
    \midrule
    S1:~~salience vocab (G-L2) & .04 & .98 & .47 & 5 \\
    S2:~~salience emb (G-L2) & .02 & .81 & .22 & 5 \\
    \bottomrule
    \end{tabular}
    \caption{Spurious Correlations in IMDB with 20 Clusters:
    \emph{Precision} and \emph{Recall} of the cluster that has the highest precision for the respective representation.
    Both are computed based on 245 numeric examples in the data.
    \emph{Size} is the data fraction of the selected cluster compared to the entire set.
    }
    \label{tab:appendix:imdb_numeric_representation_comparison_20_clusters}
\end{table}

\subsection{Spurious Correlations in IMDB}\label{sec:appendix:problematic_patterns}
\begin{figure}[t!]
    \centering
    \includegraphics[width=\columnwidth]{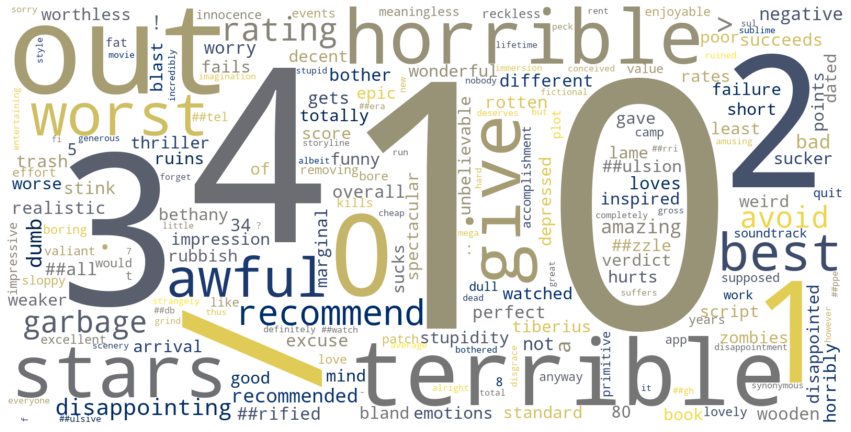}
    \caption{Spurious Correlations in IMDB: Word cloud of cluster no. 3, which has tokens `10', `4', `3', `/' and `2' as top terms.}
    \label{fig:appendix:wordcloud_imdb_numeric_cluster}
\end{figure}

Fig.~\ref{fig:appendix:wordcloud_imdb_numeric_cluster} depicts a word cloud that was computed from the numeric cluster.
Individual words are weighted by their average representation value (i.e., the mean of columns of $R$).

In the following, we list the regular expressions used to identify examples containing numeric expressions.

\begin{verbatim}
RATING_EXPRESSIONS = [
  # on, out of, ...
  r': (NUMBER) \( out of\b',
  r'\b(NUMBER) out of\b',
  r'\b(NUMBER)\* out of\b',
  r'\b(NUMBER) outta \d+\b',
  # Exclude, e.g., "my top 10 of".
  r'\b[^(top )](NUMBER) of \d+\b',
  r'\b(NUMBER) on \d+\b',
  r'\b(NUMBER) out \d+\b',
  # something / something
  r'\b(NUMBER) / 10\b',
  r'\b(NUMBER) / 5\b',
  # vote, ratings, ...
  r'\bvote \( (NUMBER) \)',
  r'\bvote (NUMBER)\b',
  r'\bvoted \" (NUMBER) \"',
  r'\bvote is (NUMBER)\b',
  r'\bgive this a (NUMBER)\b',
  r'\bgive it a (NUMBER)\b',
  r'\bgets a (NUMBER)\b',
  r'\brate this a (NUMBER)\b',
  r'\" (NUMBER) \" rating\b',
  # stars, points, ...
  r'\b(NUMBER) star\b',
  r'\b(NUMBER) - star\b',
  r'\b(NUMBER) stars\b',
  r'\b(NUMBER) points\b',
  # Exclude, e.g., "at 1 point".
  r'\b[^(at)] (NUMBER) point\b',  
]
\end{verbatim}

We substitute every \texttt{NUMBER} in these base expressions with the following expression:

\begin{verbatim}
zero|one|two|three|four|five|six|
seven|eight|nine|ten|\d+|\d+\.\d+|
\d+ \. \d+|\d+\+
\end{verbatim}

Finally, we consider an example to contain a numeric expression if at least one of the expressions is found in the text.

\subsection{Understanding Learned Patterns}\label{sec:appendix:learned_patterns}
\begin{figure}[t!]
    \centering
    \includegraphics[width=\columnwidth]{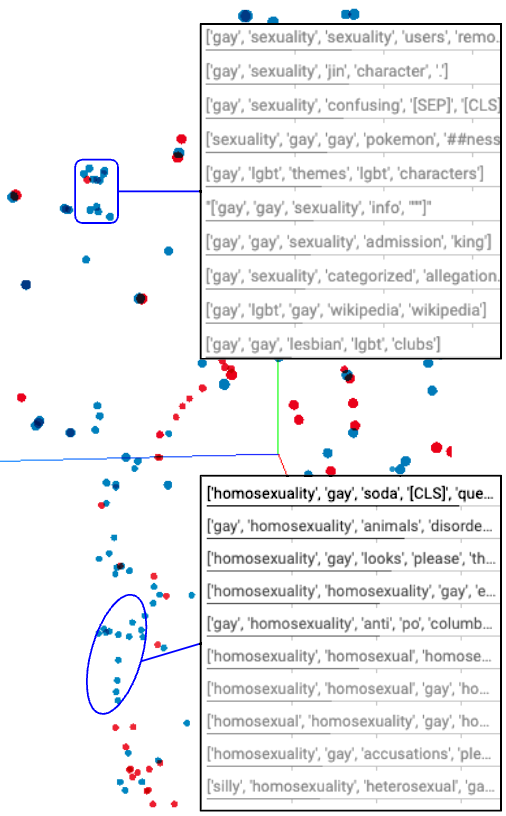}
    \caption{Patterns in Wikipedia Toxicity Subtypes: t-SNE visualization of Toxicity test examples mentioning \textit{`gay'}. Two non-toxic clusters are selected and the 10 nearest neighbors (represented here by their most salient terms in the boxes) of the respective centrally located points are shown. All these examples have \textit{`(homo)sexual(ity)'} or \textit{`LGBT'} along with \textit{`gay'} among their most salient terms.
    }
    \label{fig:appendix:gay_clusters}
\end{figure}

Fig.~\ref{fig:appendix:gay_clusters} zooms into two non-toxic clusters computed over all the test examples mentioning the term \textit{`gay'}.

\subsection{Identifying Model Sensitivities}\label{sec:appendix:model_sensitivity}
Tab.~\ref{tab:appendix:sensitivity_seed_lists} lists all seed terms we used in the sensitivity analysis of Sec.~\ref{sec:model_sensitivity}.

\begin{table*}[t]
    \centering
    \begin{tabular}{l l}
    \toprule
    Seed List & Seed Terms \\
    \midrule
    \citet{DixLiSor+18n} & atheist, queer, gay, transgender, lesbian, homosexual, feminist, black, \\
    & white, heterosexual, islam, muslim, bisexual \\
    PMI(token; toxic class) & fuck, fucking, nipple, bitch, suck, sucking, lick, cock, asshole, sucker \\
    aggregated salient terms & fuck, fucking, shit, stupid, idiot, bitch, suck, asshole, ass, dick \\
    \bottomrule
    \end{tabular}
    \caption{Identifying Model Sensitivities: Seed lists from various sources.}
    \label{tab:appendix:sensitivity_seed_lists}
\end{table*}

\subsection{Generalizing from Single Examples}\label{sec:appendix:single_example_generalization}
Tab.~\ref{tab:appendix:toxicity:examples_obvious} lists 5 examples whose predictions are not surprising.
Looking at the examples' text is enough to justify the \emph{toxic} label.

\begin{table*}[t]
\centering
\small
\begin{tabular*}{\textwidth}{p{.1cm} p{9.cm} l l}
\toprule
    & Example &  Top- Terms & Label / Prediction \\
    
    \midrule
    1. &
    failure at life loser . &
    loser, failure, life, [\textsc{sep}], [\textsc{cls}] &
    non-toxic / toxic \\
    
    2. &
    = = = = oh the irony = = = it just goes to show how stupid some americans really are , a scumbag mob lawyer mayor of las vegas . you couldn ' t make it up ! ! ! !  &
    stupid, irony, mo, \#\#um, americans &
    non-toxic / toxic \\
    
    3. &
    : oh and don ' t be a patronising ass with comments , like justin , justin . ' ' ' ' ' ' ' '  &
    ass, patron, \#\#ising, comments, with &
    non-toxic / toxic \\
    
    4. &
    what the f * ck is your problem ? get a life . you don ' t own any pages .  &
    ck, f, *, problem, pages &
    non-toxic / toxic \\
    
    5. &
    stop f * cking up the call - out tables please .  &
    f, ck, *, please, tables &
    non-toxic / toxic \\

\bottomrule
\end{tabular*}
\caption{Generalizing from Single Examples: 5 examples with their top-5 terms whose labels are not surprising.}
\label{tab:appendix:toxicity:examples_obvious}
\end{table*}

%% file: main.bbl
\begin{thebibliography}{59}
\providecommand{\natexlab}[1]{#1}

\bibitem[{Abadi et~al.(2016)Abadi, Agarwal, Barham, Brevdo, Chen, Citro,
  Corrado, Davis, Dean, Devin, Ghemawat, Goodfellow, Harp, Irving, Isard, Jia,
  J{\'{o}}zefowicz, Kaiser, Kudlur, Levenberg, Man{\'{e}}, Monga, Moore,
  Murray, Olah, Schuster, Shlens, Steiner, Sutskever, Talwar, Tucker,
  Vanhoucke, Vasudevan, Vi{\'{e}}gas, Vinyals, Warden, Wattenberg, Wicke, Yu,
  and Zheng}]{AbaAgaBar+16n}
Abadi, M.; Agarwal, A.; Barham, P.; Brevdo, E.; Chen, Z.; Citro, C.; Corrado,
  G.~S.; Davis, A.; Dean, J.; Devin, M.; Ghemawat, S.; Goodfellow, I.~J.; Harp,
  A.; Irving, G.; Isard, M.; Jia, Y.; J{\'{o}}zefowicz, R.; Kaiser, L.; Kudlur,
  M.; Levenberg, J.; Man{\'{e}}, D.; Monga, R.; Moore, S.; Murray, D.~G.; Olah,
  C.; Schuster, M.; Shlens, J.; Steiner, B.; Sutskever, I.; Talwar, K.; Tucker,
  P.~A.; Vanhoucke, V.; Vasudevan, V.; Vi{\'{e}}gas, F.~B.; Vinyals, O.;
  Warden, P.; Wattenberg, M.; Wicke, M.; Yu, Y.; and Zheng, X. 2016.
\newblock TensorFlow: Large-Scale Machine Learning on Heterogeneous Distributed
  Systems.
\newblock \emph{CoRR}, abs/1603.04467.

\bibitem[{Arras et~al.(2016)Arras, Horn, Montavon, M{\"{u}}ller, and
  Samek}]{ArrHorMon+16a}
Arras, L.; Horn, F.; Montavon, G.; M{\"{u}}ller, K.; and Samek, W. 2016.
\newblock Explaining Predictions of Non-Linear Classifiers in {NLP}.
\newblock In Blunsom, P.; Cho, K.; Cohen, S.~B.; Grefenstette, E.; Hermann,
  K.~M.; Rimell, L.; Weston, J.; and Yih, S.~W., eds., \emph{Proceedings of the
  1st Workshop on Representation Learning for NLP, Rep4NLP@ACL 2016, Berlin,
  Germany, August 11, 2016}, 1--7. Association for Computational Linguistics.

\bibitem[{Bach et~al.(2015{\natexlab{a}})Bach, Binder, Montavon, Klauschen,
  M{\"u}ller, and Samek}]{BacBinMon+15a}
Bach, S.; Binder, A.; Montavon, G.; Klauschen, F.; M{\"u}ller, K.-R.; and
  Samek, W. 2015{\natexlab{a}}.
\newblock On Pixel-Wise Explanations for Non-Linear Classifier Decisions by
  Layer-Wise Relevance Propagation.
\newblock \emph{PLOS ONE}, 10(7): 1--46.

\bibitem[{Bach et~al.(2015{\natexlab{b}})Bach, Binder, Montavon, Klauschen,
  Müller, and Samek}]{bach-etal-2015}
Bach, S.; Binder, A.; Montavon, G.; Klauschen, F.; Müller, K.-R.; and Samek,
  W. 2015{\natexlab{b}}.
\newblock On Pixel-Wise Explanations for Non-Linear Classifier Decisions by
  Layer-Wise Relevance Propagation.
\newblock \emph{PLoS ONE}, 10(7).

\bibitem[{Bahdanau, Cho, and Bengio(2015)}]{BahChoBen15z}
Bahdanau, D.; Cho, K.; and Bengio, Y. 2015.
\newblock Neural Machine Translation by Jointly Learning to Align and
  Translate.
\newblock In Bengio, Y.; and LeCun, Y., eds., \emph{3rd International
  Conference on Learning Representations, {ICLR} 2015, San Diego, CA, USA, May
  7-9, 2015, Conference Track Proceedings}.

\bibitem[{Bastings et~al.(2022)Bastings, Ebert, Zablotskaia, Sandholm, and
  Filippova}]{bastings-et-al-will-you}
Bastings, J.; Ebert, S.; Zablotskaia, P.; Sandholm, A.; and Filippova, K. 2022.
\newblock "Will {Y}ou {F}ind {T}hese {S}hortcuts?": {A} {P}rotocol for
  {E}valuating the {F}aithfulness of {I}nput {S}alience {M}ethods for {T}ext
  {C}lassification.
\newblock In \emph{Proceedings of the 2021 Conference on Empirical Methods in
  Natural Language Processing (to appear)}.

\bibitem[{Bastings and Filippova(2020)}]{bastings-filippova-2020-elephant}
Bastings, J.; and Filippova, K. 2020.
\newblock The elephant in the interpretability room: Why use attention as
  explanation when we have saliency methods?
\newblock In \emph{Proceedings of the Third BlackboxNLP Workshop on Analyzing
  and Interpreting Neural Networks for NLP}, 149--155. Online: Association for
  Computational Linguistics.

\bibitem[{Bowman et~al.(2015)Bowman, Angeli, Potts, and
  Manning}]{bowman-etal-2015-nli}
Bowman, S.~R.; Angeli, G.; Potts, C.; and Manning, C.~D. 2015.
\newblock A large annotated corpus for learning natural language inference.
\newblock In \emph{Proceedings of the 2015 Conference on Empirical Methods in
  Natural Language Processing}, 632--642. Lisbon, Portugal: Association for
  Computational Linguistics.

\bibitem[{Danilevsky et~al.(2020)Danilevsky, Qian, Aharonov, Katsis, Kawas, and
  Sen}]{danilevsky-etal-2020-survey}
Danilevsky, M.; Qian, K.; Aharonov, R.; Katsis, Y.; Kawas, B.; and Sen, P.
  2020.
\newblock A Survey of the State of Explainable {AI} for Natural Language
  Processing.
\newblock In \emph{Proceedings of the 1st Conference of the Asia-Pacific
  Chapter of the Association for Computational Linguistics and the 10th
  International Joint Conference on Natural Language Processing}, 447--459.
  Suzhou, China: Association for Computational Linguistics.

\bibitem[{Denil, Demiraj, and de~Freitas(2015)}]{denil-et-al-2015-extraction}
Denil, M.; Demiraj, A.; and de~Freitas, N. 2015.
\newblock Extraction of Salient Sentences from Labelled Documents.
\newblock arXiv:1412.6815.

\bibitem[{Devlin et~al.(2019)Devlin, Chang, Lee, and Toutanova}]{DevChaLee+19a}
Devlin, J.; Chang, M.; Lee, K.; and Toutanova, K. 2019.
\newblock {BERT:} Pre-training of Deep Bidirectional Transformers for Language
  Understanding.
\newblock In \emph{Proceedings of the 2019 Conference of the North American
  Chapter of the Association for Computational Linguistics: Human Language
  Technologies, {NAACL-HLT} 2019, Minneapolis, MN, USA, June 2-7, 2019, Volume
  1 (Long and Short Papers)}, 4171--4186.

\bibitem[{DeYoung et~al.(2020)DeYoung, Jain, Rajani, Lehman, Xiong, Socher, and
  Wallace}]{deyoung-etal-2020-eraser}
DeYoung, J.; Jain, S.; Rajani, N.~F.; Lehman, E.; Xiong, C.; Socher, R.; and
  Wallace, B.~C. 2020.
\newblock {ERASER}: {A} Benchmark to Evaluate Rationalized {NLP} Models.
\newblock In \emph{Proceedings of the 58th Annual Meeting of the Association
  for Computational Linguistics}, 4443--4458. Online: Association for
  Computational Linguistics.

\bibitem[{Dixon et~al.(2018{\natexlab{a}})Dixon, Li, Sorensen, Thain, and
  Vasserman}]{dixon-etal-toxicity-2018}
Dixon, L.; Li, J.; Sorensen, J.; Thain, N.; and Vasserman, L.
  2018{\natexlab{a}}.
\newblock Measuring and Mitigating Unintended Bias in Text Classification.
\newblock In \emph{AAAI/ACM Conference on AI, Ethics, and Society}.

\bibitem[{Dixon et~al.(2018{\natexlab{b}})Dixon, Li, Sorensen, Thain, and
  Vasserman}]{DixLiSor+18n}
Dixon, L.; Li, J.; Sorensen, J.; Thain, N.; and Vasserman, L.
  2018{\natexlab{b}}.
\newblock Measuring and Mitigating Unintended Bias in Text Classification.
\newblock In Furman, J.; Marchant, G.~E.; Price, H.; and Rossi, F., eds.,
  \emph{Proceedings of the 2018 {AAAI/ACM} Conference on AI, Ethics, and
  Society, {AIES} 2018, New Orleans, LA, USA, February 02-03, 2018}, 67--73.
  {ACM}.

\bibitem[{Eisenstein(2022)}]{eisenstein-2022-informativeness}
Eisenstein, J. 2022.
\newblock Informativeness and Invariance: Two Perspectives on Spurious
  Correlations in Natural Language.
\newblock In \emph{Proceedings of the 2022 Conference of the North American
  Chapter of the Association for Computational Linguistics: Human Language
  Technologies}, 4326--4331. Seattle, United States: Association for
  Computational Linguistics.

\bibitem[{Gardner et~al.(2021)Gardner, Merrill, Dodge, Peters, Ross, Singh, and
  Smith}]{gardner-etal-2021-competency}
Gardner, M.; Merrill, W.; Dodge, J.; Peters, M.; Ross, A.; Singh, S.; and
  Smith, N.~A. 2021.
\newblock Competency Problems: On Finding and Removing Artifacts in Language
  Data.
\newblock In \emph{Proceedings of the 2021 Conference on Empirical Methods in
  Natural Language Processing}, 1801--1813. Online and Punta Cana, Dominican
  Republic: Association for Computational Linguistics.

\bibitem[{Geirhos et~al.(2020)Geirhos, Jacobsen, Michaelis, Zemel, Brendel,
  Bethge, and Wichmann}]{geirhos-2020-shortcuts}
Geirhos, R.; Jacobsen, J.-H.; Michaelis, C.; Zemel, R.; Brendel, W.; Bethge,
  M.; and Wichmann, F.~A. 2020.
\newblock Shortcut Learning in Deep Neural Networks.
\newblock \emph{Nature Machine Intelligence}, 2: 665--–673.

\bibitem[{Geva et~al.(2022)Geva, Caciularu, Dar, Roit, Sadde, Shlain, Tamir,
  and Goldberg}]{GevCacDar+22s}
Geva, M.; Caciularu, A.; Dar, G.; Roit, P.; Sadde, S.; Shlain, M.; Tamir, B.;
  and Goldberg, Y. 2022.
\newblock LM-Debugger: An Interactive Tool for Inspection and Intervention in
  Transformer-Based Language Models.
\newblock \emph{CoRR}, abs/2204.12130.

\bibitem[{Geva, Goldberg, and Berant(2019)}]{geva-etal-2019-modeling}
Geva, M.; Goldberg, Y.; and Berant, J. 2019.
\newblock Are We Modeling the Task or the Annotator? An Investigation of
  Annotator Bias in Natural Language Understanding Datasets.
\newblock In \emph{Proceedings of the 2019 Conference on Empirical Methods in
  Natural Language Processing and the 9th International Joint Conference on
  Natural Language Processing (EMNLP-IJCNLP)}, 1161--1166. Hong Kong, China:
  Association for Computational Linguistics.

\bibitem[{Gururangan et~al.(2018)Gururangan, Swayamdipta, Levy, Schwartz,
  Bowman, and Smith}]{gururangan-etal-2018-annotation}
Gururangan, S.; Swayamdipta, S.; Levy, O.; Schwartz, R.; Bowman, S.; and Smith,
  N.~A. 2018.
\newblock Annotation Artifacts in Natural Language Inference Data.
\newblock In \emph{Proceedings of the 2018 Conference of the North {A}merican
  Chapter of the Association for Computational Linguistics: Human Language
  Technologies, Volume 2 (Short Papers)}, 107--112. New Orleans, Louisiana:
  Association for Computational Linguistics.

\bibitem[{Han and Tsvetkov(2021)}]{han-tsvetkov-2021-influence-tuning}
Han, X.; and Tsvetkov, Y. 2021.
\newblock Influence Tuning: Demoting Spurious Correlations via Instance
  Attribution and Instance-Driven Updates.
\newblock In \emph{Findings of the Association for Computational Linguistics:
  EMNLP 2021}, 4398--4409. Punta Cana, Dominican Republic: Association for
  Computational Linguistics.

\bibitem[{Hartvigsen et~al.(2022)Hartvigsen, Gabriel, Palangi, Sap, Ray, and
  Kamar}]{hartvigsen-etal-2022-toxigen}
Hartvigsen, T.; Gabriel, S.; Palangi, H.; Sap, M.; Ray, D.; and Kamar, E. 2022.
\newblock {T}oxi{G}en: A Large-Scale Machine-Generated Dataset for Adversarial
  and Implicit Hate Speech Detection.
\newblock In \emph{Proceedings of the 60th Annual Meeting of the Association
  for Computational Linguistics (Volume 1: Long Papers)}, 3309--3326. Dublin,
  Ireland: Association for Computational Linguistics.

\bibitem[{Kauffmann et~al.(2022)Kauffmann, Esders, Ruff, Montavon, Samek, and
  M{\"u}ller}]{KauEsdRuf+22a}
Kauffmann, J.; Esders, M.; Ruff, L.; Montavon, G.; Samek, W.; and M{\"u}ller,
  K.-R. 2022.
\newblock From Clustering to Cluster Explanations via Neural Networks.
\newblock \emph{IEEE Transactions on Neural Networks and Learning Systems},
  1--15.

\bibitem[{Kaushik, Hovy, and Lipton(2020)}]{kaushik-etal-2020-learning}
Kaushik, D.; Hovy, E.~H.; and Lipton, Z.~C. 2020.
\newblock Learning The Difference That Makes A Difference With
  Counterfactually-Augmented Data.
\newblock In \emph{8th International Conference on Learning Representations,
  ICLR 2020, Addis Ababa, Ethiopia, April 26-30, 2020}. OpenReview.net.

\bibitem[{Kingma and Ba(2015)}]{KinBa15s}
Kingma, D.~P.; and Ba, J. 2015.
\newblock Adam: {A} Method for Stochastic Optimization.
\newblock In Bengio, Y.; and LeCun, Y., eds., \emph{3rd International
  Conference on Learning Representations, {ICLR} 2015, San Diego, CA, USA, May
  7-9, 2015, Conference Track Proceedings}.

\bibitem[{Koh and Liang(2017)}]{koh-liang-if-2017}
Koh, P.~W.; and Liang, P. 2017.
\newblock Understanding Black-box Predictions via Influence Functions.
\newblock In Precup, D.; and Teh, Y.~W., eds., \emph{Proceedings of the 34th
  International Conference on Machine Learning}, volume~70 of \emph{Proceedings
  of Machine Learning Research}, 1885--1894. PMLR.

\bibitem[{Kokhlikyan et~al.(2020)Kokhlikyan, Miglani, Martin, Wang, Alsallakh,
  Reynolds, Melnikov, Kliushkina, Araya, Yan, and
  Reblitz{-}Richardson}]{KokMigMar+20o}
Kokhlikyan, N.; Miglani, V.; Martin, M.; Wang, E.; Alsallakh, B.; Reynolds, J.;
  Melnikov, A.; Kliushkina, N.; Araya, C.; Yan, S.; and Reblitz{-}Richardson,
  O. 2020.
\newblock Captum: {A} unified and generic model interpretability library for
  PyTorch.
\newblock \emph{CoRR}, abs/2009.07896.

\bibitem[{Lapuschkin et~al.(2019)Lapuschkin, W{\"a}ldchen, Binder, Montavon,
  Samek, and M{\"u}ller}]{lapuschkin-2019-spray}
Lapuschkin, S.; W{\"a}ldchen, S.; Binder, A.; Montavon, G.; Samek, W.; and
  M{\"u}ller, K.-R. 2019.
\newblock Unmasking Clever Hans Predictors and Assessing What Machines Really
  Learn.
\newblock \emph{Nature Communications}, 10: 1096.

\bibitem[{Lertvittayakumjorn, Specia, and Toni(2020)}]{LerSpeTon20b}
Lertvittayakumjorn, P.; Specia, L.; and Toni, F. 2020.
\newblock {FIND:} Human-in-the-Loop Debugging Deep Text Classifiers.
\newblock In Webber, B.; Cohn, T.; He, Y.; and Liu, Y., eds., \emph{Proceedings
  of the 2020 Conference on Empirical Methods in Natural Language Processing,
  {EMNLP} 2020, Online, November 16-20, 2020}, 332--348. Association for
  Computational Linguistics.

\bibitem[{Lertvittayakumjorn and
  Toni(2021{\natexlab{a}})}]{lertvittayakumjorn-toni-2021-explanation}
Lertvittayakumjorn, P.; and Toni, F. 2021{\natexlab{a}}.
\newblock Explanation-Based Human Debugging of {NLP} Models: A Survey.
\newblock \emph{Transactions of the Association for Computational Linguistics},
  9: 1508--1528.

\bibitem[{Lertvittayakumjorn and Toni(2021{\natexlab{b}})}]{LerTon21a}
Lertvittayakumjorn, P.; and Toni, F. 2021{\natexlab{b}}.
\newblock Explanation-Based Human Debugging of {NLP} Models: {A} Survey.
\newblock \emph{Trans. Assoc. Comput. Linguistics}, 9: 1508--1528.

\bibitem[{Li et~al.(2016)Li, Chen, Hovy, and
  Jurafsky}]{li-etal-2016-visualizing}
Li, J.; Chen, X.; Hovy, E.; and Jurafsky, D. 2016.
\newblock Visualizing and Understanding Neural Models in {NLP}.
\newblock In \emph{Proceedings of the 2016 Conference of the North {A}merican
  Chapter of the Association for Computational Linguistics: Human Language
  Technologies}, 681--691. San Diego, California: Association for Computational
  Linguistics.

\bibitem[{Li, Monroe, and Jurafsky(2016)}]{li-etal-2016-understanding}
Li, J.; Monroe, W.; and Jurafsky, D. 2016.
\newblock Understanding Neural Networks through Representation Erasure.
\newblock arXiv:1612.08220.

\bibitem[{Maas et~al.(2011)Maas, Daly, Pham, Huang, Ng, and
  Potts}]{maas-etal-2011-imdb}
Maas, A.~L.; Daly, R.~E.; Pham, P.~T.; Huang, D.; Ng, A.~Y.; and Potts, C.
  2011.
\newblock Learning Word Vectors for Sentiment Analysis.
\newblock In \emph{Proceedings of the 49th Annual Meeting of the Association
  for Computational Linguistics: Human Language Technologies}, 142--150.
  Portland, Oregon, USA: Association for Computational Linguistics.

\bibitem[{McCoy, Pavlick, and Linzen(2019)}]{mccoy-etal-2019-right}
McCoy, T.; Pavlick, E.; and Linzen, T. 2019.
\newblock Right for the Wrong Reasons: Diagnosing Syntactic Heuristics in
  Natural Language Inference.
\newblock In \emph{Proceedings of the 57th Annual Meeting of the Association
  for Computational Linguistics}, 3428--3448. Florence, Italy: Association for
  Computational Linguistics.

\bibitem[{Mitchell et~al.(2019)Mitchell, Wu, Zaldivar, Barnes, Vasserman,
  Hutchinson, Spitzer, Raji, and Gebru}]{MitWuZal+19j}
Mitchell, M.; Wu, S.; Zaldivar, A.; Barnes, P.; Vasserman, L.; Hutchinson, B.;
  Spitzer, E.; Raji, I.~D.; and Gebru, T. 2019.
\newblock Model Cards for Model Reporting.
\newblock In danah boyd; and Morgenstern, J.~H., eds., \emph{Proceedings of the
  Conference on Fairness, Accountability, and Transparency, FAT* 2019, Atlanta,
  GA, USA, January 29-31, 2019}, 220--229. {ACM}.

\bibitem[{Montavon et~al.(2019)Montavon, Binder, Lapuschkin, Samek, and
  M{\"{u}}ller}]{montavon-lrp-overview-2019}
Montavon, G.; Binder, A.; Lapuschkin, S.; Samek, W.; and M{\"{u}}ller, K. 2019.
\newblock Layer-Wise Relevance Propagation: An Overview.
\newblock In Samek, W.; Montavon, G.; Vedaldi, A.; Hansen, L.~K.; and
  M{\"{u}}ller, K., eds., \emph{Explainable {AI:} Interpreting, Explaining and
  Visualizing Deep Learning}, volume 11700 of \emph{Lecture Notes in Computer
  Science}, 193--209. Springer.

\bibitem[{Nori et~al.(2019)Nori, Jenkins, Koch, and Caruana}]{NorJenKoc+19v}
Nori, H.; Jenkins, S.; Koch, P.; and Caruana, R. 2019.
\newblock InterpretML: A Unified Framework for Machine Learning
  Interpretability.
\newblock \emph{arXiv preprint arXiv:1909.09223}.

\bibitem[{Pedregosa et~al.(2011)Pedregosa, Varoquaux, Gramfort, Michel,
  Thirion, Grisel, Blondel, Prettenhofer, Weiss, Dubourg, VanderPlas, Passos,
  Cournapeau, Brucher, Perrot, and Duchesnay}]{PedVarGra+11v}
Pedregosa, F.; Varoquaux, G.; Gramfort, A.; Michel, V.; Thirion, B.; Grisel,
  O.; Blondel, M.; Prettenhofer, P.; Weiss, R.; Dubourg, V.; VanderPlas, J.;
  Passos, A.; Cournapeau, D.; Brucher, M.; Perrot, M.; and Duchesnay, E. 2011.
\newblock Scikit-learn: Machine Learning in Python.
\newblock \emph{J. Mach. Learn. Res.}, 12: 2825--2830.

\bibitem[{Pezeshkpour et~al.(2021)Pezeshkpour, Jain, Singh, and
  Wallace}]{pezeshkpour-etal-2021-combining}
Pezeshkpour, P.; Jain, S.; Singh, S.; and Wallace, B.~C. 2021.
\newblock Combining Feature and Instance Attribution to Detect Artifacts.
\newblock arXiv:2107.00323.

\bibitem[{Pruthi et~al.(2020)Pruthi, Liu, Kale, and
  Sundararajan}]{pruthi-tracin-20}
Pruthi, G.; Liu, F.; Kale, S.; and Sundararajan, M. 2020.
\newblock Estimating Training Data Influence by Tracing Gradient Descent.
\newblock In Larochelle, H.; Ranzato, M.; Hadsell, R.; Balcan, M.-F.; and Lin,
  H.-T., eds., \emph{Advances in Neural Information Processing Systems 33:
  Annual Conference on Neural Information Processing Systems 2020, NeurIPS
  2020, December 6-12, 2020, virtual}.

\bibitem[{Pushkarna, Zaldivar, and Kjartansson(2022)}]{PusZalKja22l}
Pushkarna, M.; Zaldivar, A.; and Kjartansson, O. 2022.
\newblock Data Cards: Purposeful and Transparent Dataset Documentation for
  Responsible {AI}.
\newblock In \emph{FAccT '22: 2022 {ACM} Conference on Fairness,
  Accountability, and Transparency, Seoul, Republic of Korea, June 21 - 24,
  2022}, 1776--1826. {ACM}.

\bibitem[{Ribeiro, Singh, and Guestrin(2016)}]{ribeiro-lime}
Ribeiro, M.~T.; Singh, S.; and Guestrin, C. 2016.
\newblock "Why Should {I} Trust You?": Explaining the Predictions of Any
  Classifier.
\newblock In \emph{Proceedings of the 22nd {ACM} {SIGKDD} International
  Conference on Knowledge Discovery and Data Mining, San Francisco, CA, USA,
  August 13-17, 2016}, 1135--1144.

\bibitem[{Ribeiro et~al.(2020)Ribeiro, Wu, Guestrin, and
  Singh}]{ribeiro-etal-2020-beyond}
Ribeiro, M.~T.; Wu, T.; Guestrin, C.; and Singh, S. 2020.
\newblock Beyond Accuracy: Behavioral Testing of {NLP} Models with
  {C}heck{L}ist.
\newblock In \emph{Proceedings of the 58th Annual Meeting of the Association
  for Computational Linguistics}, 4902--4912. Online: Association for
  Computational Linguistics.

\bibitem[{Rosenman, Jacovi, and Goldberg(2020)}]{rosenman-etal-2020-exposing}
Rosenman, S.; Jacovi, A.; and Goldberg, Y. 2020.
\newblock {E}xposing {S}hallow {H}euristics of {R}elation {E}xtraction {M}odels
  with {C}hallenge {D}ata.
\newblock In \emph{Proceedings of the 2020 Conference on Empirical Methods in
  Natural Language Processing (EMNLP)}, 3702--3710. Online: Association for
  Computational Linguistics.

\bibitem[{Ross, Marasovic, and Peters(2021)}]{ross-etal-2021-explaining}
Ross, A.; Marasovic, A.; and Peters, M.~E. 2021.
\newblock Explaining {NLP} Models via Minimal Contrastive Editing (MiCE).
\newblock In Zong, C.; Xia, F.; Li, W.; and Navigli, R., eds., \emph{Findings
  of the Association for Computational Linguistics: {ACL/IJCNLP} 2021, Online
  Event, August 1-6, 2021}, volume {ACL/IJCNLP} 2021 of \emph{Findings of
  {ACL}}, 3840--3852. Association for Computational Linguistics.

\bibitem[{Sanh et~al.(2019)Sanh, Debut, Chaumond, and Wolf}]{SanDebCha+19h}
Sanh, V.; Debut, L.; Chaumond, J.; and Wolf, T. 2019.
\newblock DistilBERT, a distilled version of {BERT:} smaller, faster, cheaper
  and lighter.
\newblock \emph{CoRR}, abs/1910.01108.

\bibitem[{Schoop et~al.(2022)Schoop, Wedin, Kapishnikov, Bolukbasi, and
  Terry}]{schoop-etal-2022-imacs}
Schoop, E.; Wedin, B.; Kapishnikov, A.; Bolukbasi, T.; and Terry, M. 2022.
\newblock IMACS: Image Model Attribution Comparison Summaries.

\bibitem[{Socher et~al.(2013)Socher, Perelygin, Wu, Chuang, Manning, Ng, and
  Potts}]{socher-etal-2013-recursive}
Socher, R.; Perelygin, A.; Wu, J.; Chuang, J.; Manning, C.~D.; Ng, A.; and
  Potts, C. 2013.
\newblock Recursive Deep Models for Semantic Compositionality Over a Sentiment
  Treebank.
\newblock In \emph{Proceedings of the 2013 Conference on Empirical Methods in
  Natural Language Processing}, 1631--1642. Seattle, Washington, USA:
  Association for Computational Linguistics.

\bibitem[{Sundararajan, Taly, and Yan(2017)}]{sundararajan-ig-2017}
Sundararajan, M.; Taly, A.; and Yan, Q. 2017.
\newblock Axiomatic Attribution for Deep Networks.
\newblock In Precup, D.; and Teh, Y.~W., eds., \emph{Proceedings of the 34th
  International Conference on Machine Learning, {ICML} 2017, Sydney, NSW,
  Australia, 6-11 August 2017}, volume~70 of \emph{Proceedings of Machine
  Learning Research}, 3319--3328. {PMLR}.

\bibitem[{Tenney et~al.(2020)Tenney, Wexler, Bastings, Bolukbasi, Coenen,
  Gehrmann, Jiang, Pushkarna, Radebaugh, Reif, and
  Yuan}]{tenney-etal-2020-language}
Tenney, I.; Wexler, J.; Bastings, J.; Bolukbasi, T.; Coenen, A.; Gehrmann, S.;
  Jiang, E.; Pushkarna, M.; Radebaugh, C.; Reif, E.; and Yuan, A. 2020.
\newblock The Language Interpretability Tool: Extensible, Interactive
  Visualizations and Analysis for {NLP} Models.
\newblock In \emph{Proceedings of the 2020 Conference on Empirical Methods in
  Natural Language Processing: System Demonstrations}, 107--118. Online:
  Association for Computational Linguistics.

\bibitem[{Van~der Maaten and Hinton(2008)}]{VanHin08m}
Van~der Maaten, L.; and Hinton, G. 2008.
\newblock Visualizing data using t-SNE.
\newblock \emph{Journal of machine learning research}, 9(11).

\bibitem[{Vaswani et~al.(2017)Vaswani, Shazeer, Parmar, Uszkoreit, Jones,
  Gomez, Kaiser, and Polosukhin}]{VasShaPar+17a}
Vaswani, A.; Shazeer, N.; Parmar, N.; Uszkoreit, J.; Jones, L.; Gomez, A.~N.;
  Kaiser, L.; and Polosukhin, I. 2017.
\newblock Attention Is All You Need.

\bibitem[{Wallace et~al.(2019)Wallace, Tuyls, Wang, Subramanian, Gardner, and
  Singh}]{WalTuyWan+19o}
Wallace, E.; Tuyls, J.; Wang, J.; Subramanian, S.; Gardner, M.; and Singh, S.
  2019.
\newblock AllenNLP Interpret: {A} Framework for Explaining Predictions of {NLP}
  Models.
\newblock In Pad{\'{o}}, S.; and Huang, R., eds., \emph{Proceedings of the 2019
  Conference on Empirical Methods in Natural Language Processing and the 9th
  International Joint Conference on Natural Language Processing, {EMNLP-IJCNLP}
  2019, Hong Kong, China, November 3-7, 2019 - System Demonstrations}, 7--12.
  Association for Computational Linguistics.

\bibitem[{Wong, Paritosh, and Bollacker(2022)}]{wong-etal-2022-reproducible}
Wong, K.; Paritosh, P.; and Bollacker, K. 2022.
\newblock Are Ground Truth Labels Reproducible? An Empirical Study.
\newblock In \emph{ML Evaluation Standards (ICLR 2022 Workshop)}.

\bibitem[{Wulczyn, Thain, and Dixon(2017)}]{wiki-toxicity-data}
Wulczyn, E.; Thain, N.; and Dixon, L. 2017.
\newblock Ex Machina: Personal Attacks Seen at Scale.
\newblock In \emph{Proceedings of the 26th International Conference on World
  Wide Web}, WWW '17, 1391--1399. Republic and Canton of Geneva, CHE:
  International World Wide Web Conferences Steering Committee.
\newblock ISBN 9781450349130.

\bibitem[{Yin and Neubig(2022)}]{yin-neubig-2022-interpreting}
Yin, K.; and Neubig, G. 2022.
\newblock Interpreting Language Models with Contrastive Explanations.

\bibitem[{Zeiler and Fergus(2014)}]{zeiler-fergus-2014-visualizing}
Zeiler, M.~D.; and Fergus, R. 2014.
\newblock Visualizing and Understanding Convolutional Networks.
\newblock In Fleet, D.; Pajdla, T.; Schiele, B.; and Tuytelaars, T., eds.,
  \emph{Computer Vision -- ECCV 2014}, 818--833. Cham: Springer International
  Publishing.
\newblock ISBN 978-3-319-10590-1.

\bibitem[{Zylberajch, Lertvittayakumjorn, and
  Toni(2021)}]{zylberajch-etal-2021-hildif}
Zylberajch, H.; Lertvittayakumjorn, P.; and Toni, F. 2021.
\newblock HILDIF: interactive debugging of NLI models using influence
  functions.
\newblock 1--6. ASSOC COMPUTATIONAL LINGUISTICS-ACL.

\end{thebibliography}
